\documentclass{article} %
\usepackage{iclr2027_conference,times}

\usepackage{amsmath,amsfonts,bm}

\def\eqref#1{equation~\ref{#1}}

\def\1{\bm{1}}

\DeclareMathAlphabet{\mathsfit}{\encodingdefault}{\sfdefault}{m}{sl}
\SetMathAlphabet{\mathsfit}{bold}{\encodingdefault}{\sfdefault}{bx}{n}

\usepackage{hyperref}
\usepackage{url}
\usepackage{booktabs}       %
\usepackage{amsfonts}       %
\usepackage{nicefrac}       %
\usepackage{microtype}      %
\usepackage[table]{xcolor}         %
\usepackage{amsmath}
\usepackage{caption}
\usepackage{tabularray}
\usepackage{makecell}
\usepackage{tabularx}
\usepackage{array}
\usepackage{graphicx}
\usepackage{wrapfig}
\usepackage{subcaption}
\usepackage{enumitem}
\usepackage{multirow}

\definecolor{tablegray}{RGB}{246,247,249}
\newcolumntype{Y}{>{\centering\arraybackslash}X}

\title{AdaVSkip: Adaptive Visual Token Skipping Across Layers For Efficient MLLMs Inference}

\author{%
  Yuyao Sun\textsuperscript{1}\thanks{Equal contribution.
  \quad $^{\dagger}$Corresponding authors.}\enspace
  Tao Deng\textsuperscript{1}\footnotemark[1]\enspace
  Shuang Li\textsuperscript{1}$^{\dagger}$\enspace
  Deqing Wang\textsuperscript{1}$^{\dagger}$ \\
  \textsuperscript{1}Beihang University
}

\iclrfinalcopy
\begin{document}

\maketitle
\fancyhead{}
\renewcommand{\headrulewidth}{0pt}

\begin{abstract}
Multimodal large language models (MLLMs) require substantial computation to process numerous visual tokens across all transformer layers.
Most methods for efficient MLLM inference exploit \emph{horizontal redundancy} by compressing visual tokens. Beyond token reduction, recent studies exploit \emph{vertical redundancy} through early exit or fixed-layer skipping.
However, we find that the extent and distribution of this redundancy vary across inputs and differ between self-attention and MLP modules.
Motivated by these observations, we propose AdaVSkip, which equips each layer with two lightweight routers that independently determine whether visual tokens pass through by or skip the self-attention and MLP modules.
These decisions collectively define an input-specific visual-computation path, but their discrete and non-differentiable nature makes learning effective paths challenging.
To address this challenge, we develop a progressive two-stage training framework that updates only the routers while keeping the backbone frozen.
Stage I establishes an initial routing policy through supervised training with input-specific targets derived from module-wise necessity scores.
To further align the routing policy with task performance, Stage II uses reinforcement learning to optimize routing decisions with direct feedback from generated answers. It combines an answer correctness reward with a skip-consistency reward that discourages excessive retention of visual-token computation.
Across three MLLM backbones, AdaVSkip maintains strong task performance with substantially less computation. On LLaVA-NeXT-7B, AdaVSkip reduces FLOPs by 53.2\% while preserving the original model's average performance. Combining it with visual token compression increases this reduction to 91.2\%, while retaining 97.2\% of the original performance on average.  

\end{abstract}

\section{Introduction}

Multimodal large language models (MLLMs) have achieved remarkable progress across diverse vision-language tasks~\cite{flamingo,blip2,instructblip,llava,llava_1_5}.
Recent models increasingly adopt high-resolution inputs or fine-grained visual representations, often producing large numbers of visual tokens~\cite{llavanext,llavaonevision,internvl_1_5,internvl3,qwen3vl}.
These visual tokens are processed alongside text tokens by both self-attention and MLP modules at every transformer layer, resulting in substantial inference costs.


To improve efficiency, existing methods reduce visual-token counts by learning compact representations before feeding them into the language model~\cite{llavamini,tokenpacker,internvlx} or by pruning less informative tokens during inference~\cite{fastv,fastvlm,visionzip,fitprune,cdpruner}.
These methods primarily exploit \emph{horizontal redundancy} within the visual-token sequence, but the retained tokens may still undergo computation in layers where their contribution is limited.
To exploit this \emph{vertical redundancy} across model depth, recent studies allow visual tokens to exit early~\cite{vtw,dyvte} or freeze visual-token updates in input-invariant layer subsets~\cite{shortv}. 
However, it remains unclear whether fixed or prefix-based skipping patterns can adequately accommodate the varying visual-computation needs of different questions.


To investigate this question, we systematically analyze how skipping visual-token computation in individual self-attention and MLP modules affects the model's output distribution across different inputs.
Our analysis yields two key observations.
First, vertical redundancy is input-dependent: both the extent of skippable visual-token computation and its distribution across model depth vary substantially across inputs, limiting the effectiveness of a fixed skipping pattern.
Second, even within the same layer, self-attention and MLP often differ in their need for visual-token computation, suggesting that a shared layer-wise skipping decision may skip useful computation in one module or retain redundant computation in the other.
Together, these observations motivate an input-adaptive, module-wise visual-token skipping mechanism.


\begin{wrapfigure}{r}{0.48\linewidth}
    \centering
    \vspace{-1.5em}
    \includegraphics[width=\linewidth]{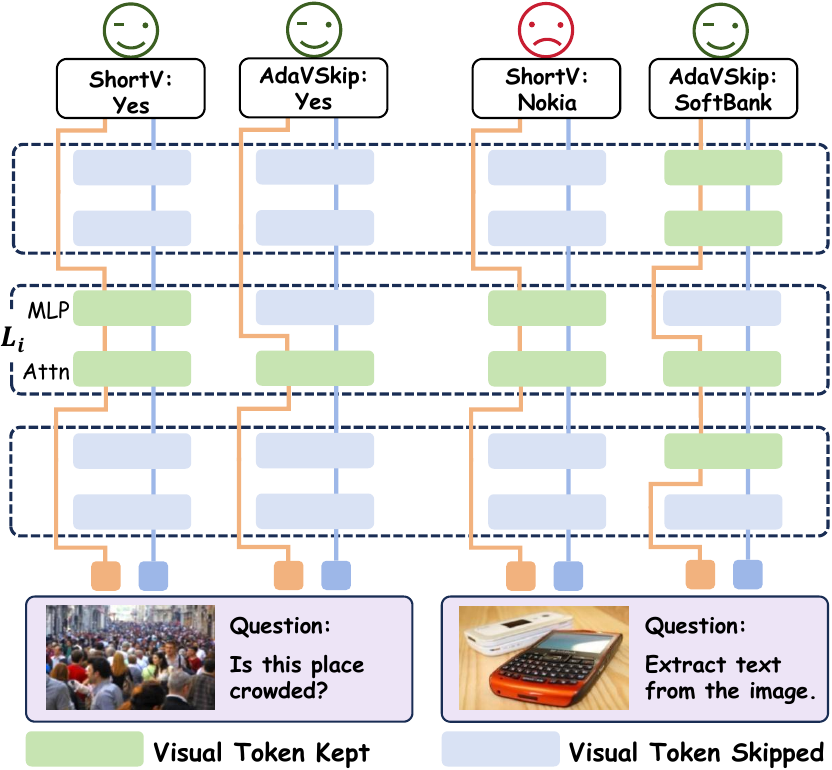}
    \caption{Fixed layer-wise skipping vs. adaptive module-wise visual token skipping.}
    \vspace{-1em}
    \label{fig:wrap}
\end{wrapfigure}

To this end, we propose \textbf{AdaVSkip}, an input-adaptive framework for module-wise visual-token skipping.
At each transformer layer, AdaVSkip introduces two lightweight routers that independently determine whether to perform visual-token computation in the self-attention and MLP modules, respectively.
When a router chooses to skip, visual tokens bypass the module with their hidden states unchanged.
They remain in the sequence and can resume computation in subsequent modules.
Text tokens continue to be processed in every module regardless of the routing decisions.
Collectively, these decisions define an input-specific visual-computation path, allowing AdaVSkip to adapt both the extent and distribution of visual-token skipping across layers and modules.


However, the non-differentiability of discrete routing decisions poses a key challenge to training these routers.
We address this challenge through a progressive two-stage training strategy that updates only the routers while keeping the backbone frozen.
In Stage I, we construct input-specific routing targets from module-wise necessity scores.
For each module type, we assign keep targets to the smallest subset of highest-scoring modules whose cumulative necessity reaches a predefined fraction of the total necessity mass.
We then train the routers using both necessity-guided supervision and the downstream language modeling loss to establish an initial routing policy.
However, this training relies on estimated routing targets and approximate gradients through discrete routing decisions, which may limit the quality of the learned policy.
Stage II therefore refines this policy through reinforcement learning, using the correctness of answers generated under sampled routing paths as direct feedback.
To discourage overly conservative retention of visual-token computation, we additionally introduce a skip-consistency reward that regularizes sampled keep ratios toward their sample-specific Stage-I references, encouraging the policy to explore better-performing routing paths at comparable computational cost. 

Extensive experiments across three MLLM backbones and multiple benchmarks demonstrate that AdaVSkip improves performance--efficiency trade-offs over fixed-layer and prefix-based skipping methods, particularly under aggressive skipping.
At moderate skipping levels, AdaVSkip reduces FLOPs by approximately 50\% while largely preserving average performance.
Moreover, AdaVSkip is complementary to visual-token compression. On LLaVA-NeXT-7B, combining it with CDPruner reduces FLOPs by 91.2\% while retaining 97.2\% of the original performance on average.

Our main contributions are summarized as follows:
\begin{itemize}
    \item We characterize vertical redundancy in visual-token computation, revealing that its extent and distribution vary across inputs and that self-attention and MLP exhibit distinct patterns.
    
    \item We propose \textbf{AdaVSkip}, an input-adaptive, module-wise visual-token skipping framework with two lightweight routers per layer that independently determine whether visual tokens participate in the computation of the self-attention and MLP modules.
    
    \item We develop a progressive two-stage training strategy that first learns an initial routing policy through necessity-guided supervision and then refines it through reinforcement learning with correctness and skip-consistency rewards, while keeping the backbone frozen.

    \item Across three MLLMs and multiple benchmarks, AdaVSkip offers better performance--efficiency trade-offs than fixed-layer and early-exit methods. With visual token compression, it saves more computation with less performance loss than token compression alone.
\end{itemize}

\vspace{-1ex}

\section{Related Work}
\subsection{Multimodal Large Language Models}
Recent multimodal large language models (MLLMs) extend the reasoning and instruction-following capabilities of LLMs to visual inputs~\cite{flamingo,blip2,instructblip,internvl_1_5,qwen2.5vl}. A widely adopted design is the projector-based paradigm, which connects a pretrained vision encoder to an LLM through a lightweight projector, mapping images into visual tokens that are processed together with text tokens in the LLM~\cite{llava,llava_1_5}. Building on this paradigm, recent MLLMs improve visual perception by using higher-resolution inputs or introducing more visual tokens for finer-grained visual representation~\cite{llavanext,internvl3,seedvl,qwen3vl}. However, this trend substantially increases computational costs, making efficient visual-token computation increasingly important for MLLMs.

\subsection{Horizontal Redundancy in Visual Tokens}

Existing methods for efficient MLLM inference mainly exploit horizontal redundancy by reducing the number of visual tokens processed by the language model.
These methods fall into two categories.
The first learns compact visual representations before the LLM stage, often through redesigned visual projectors~\cite{llavamini,tokenpacker,internvlx}, but usually requires architectural modifications and training.
The second performs training-free token reduction during inference by estimating token importance or redundancy and pruning or merging less informative tokens~\cite{fastv,sparsevlm,visionzip,fitprune,cdpruner}.
Although effective, both approaches reduce the visual-token count while processing the remaining tokens through all LLM layers, leaving vertical redundancy in visual-token computation largely unexploited.

\subsection{Vertical Redundancy Across Layers}

Vertical redundancy has been explored in LLMs through early-exit and adaptive-depth methods, including SkipDecode~\cite{skipdecode}, LayerSkip~\cite{layerskip}, Mixture-of-Depths~\cite{mixture}, and SkipGPT~\cite{skipgpt}.
In MLLMs, visual-token computation may be necessary in some layers but contribute little in others.
This vertical redundancy remains less explored than horizontal redundancy.
VTW~\cite{vtw} and DyVTE~\cite{dyvte} reduce visual-token computation through early exit, with DyVTE supporting input-dependent exit depths.
Both methods restrict visual-token computation to a prefix of layers.
Challenging this prefix assumption, ShortV~\cite{shortv} show that important visual-token computation is not confined to early layers and preserve it in selected layers.
Rather than using a fixed layer subset or a single exit point, AdaVSkip determines how much visual-token computation to skip and where for each input, with independent decisions for self-attention and MLP at each layer.

\vspace{-1ex}

\section{Method}\label{sec:method}
\subsection{Patterns of Vertical Redundancy}

To characterize vertical redundancy, we quantify the necessity of visual-token computation in the self-attention and MLP modules of each transformer layer, yielding $2L$ scores for an $L$-layer MLLM.
For each input $x$ with ground-truth answer $y=(y_1,\ldots,y_T)$, we first run the original model on the input and answer in a single forward pass with teacher forcing, recording its output distributions at all answer-token positions.
We then repeat the forward pass, skipping visual-token computation in one module at a time.
For layer $l$ and module type $c\in\mathcal{C}=\{\mathrm{attn},\mathrm{mlp}\}$, we define the necessity score by comparing the resulting distributions with those of the original model:
\begin{equation}
s_l^{c}(x)
=
\frac{1}{T}
\sum_{t=1}^{T}
D_{\mathrm{KL}}
\left(
P_{\mathrm{full}}(\cdot \mid x,y_{<t})
\,\middle\|\,
P_{-(l,c)}(\cdot \mid x,y_{<t})
\right),
\end{equation}
where $P_{\mathrm{full}}$ denotes the original model, and $P_{-(l,c)}$ denotes the intervened model where visual-token computation in the module of type $c$ at layer $l$ is skipped while other computations remain unchanged.
A larger $s_l^{c}(x)$ indicates that skipping visual-token computation in this module causes a larger deviation from the original model, suggesting higher visual-token necessity. 

\begin{figure}[t]
    \centering
    \vspace{-1em}
    \newlength{\subfigheight}
    \setlength{\subfigheight}{0.2375\textwidth}

    \subcaptionbox{Self-attn necessity across layers.\label{fig:adaptive_patterns_attn}}{
        \includegraphics[height=\subfigheight]{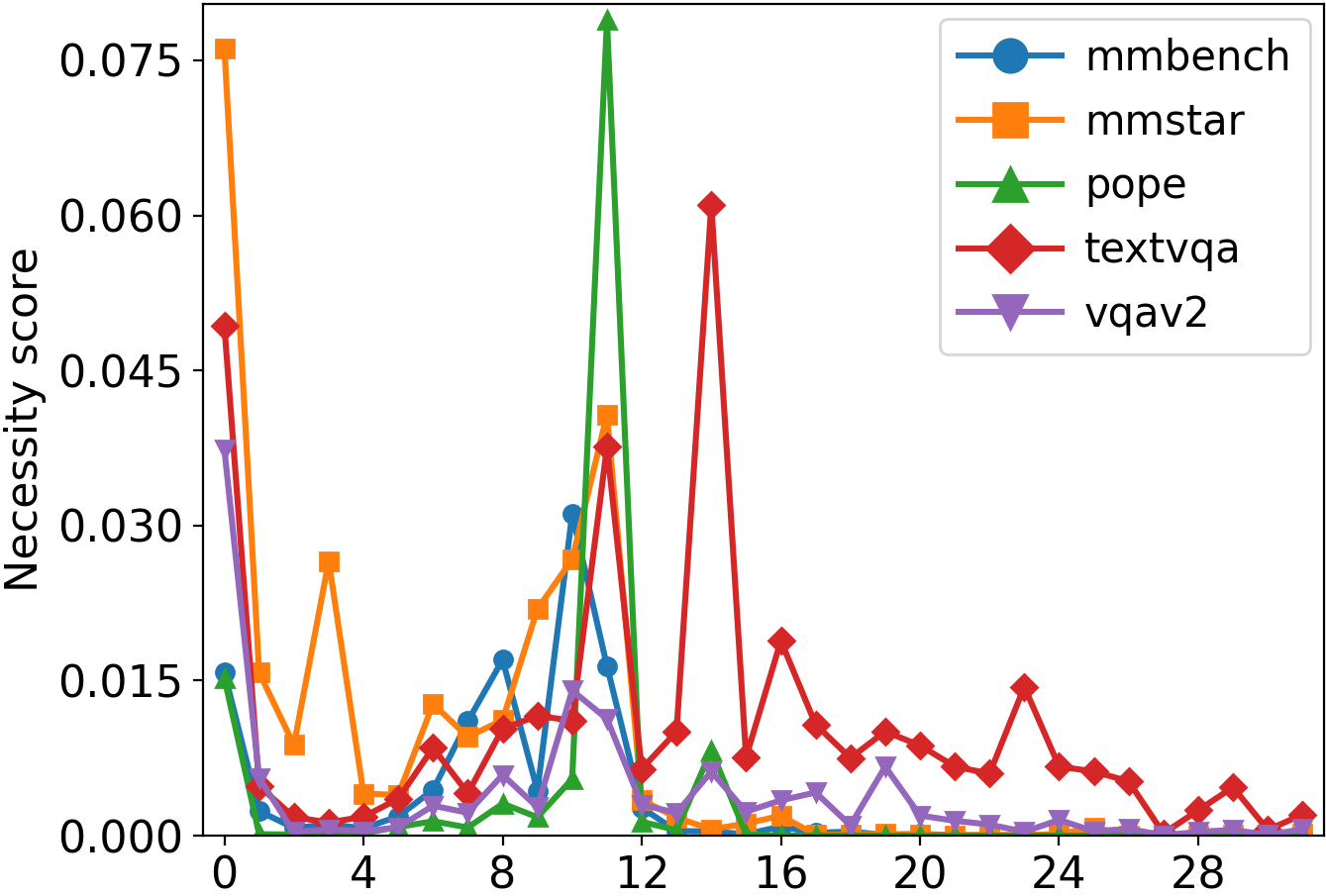}
    }
    \hfill
    \subcaptionbox{MLP necessity across layers.\label{fig:adaptive_patterns_mlp}}{
        \includegraphics[height=\subfigheight]{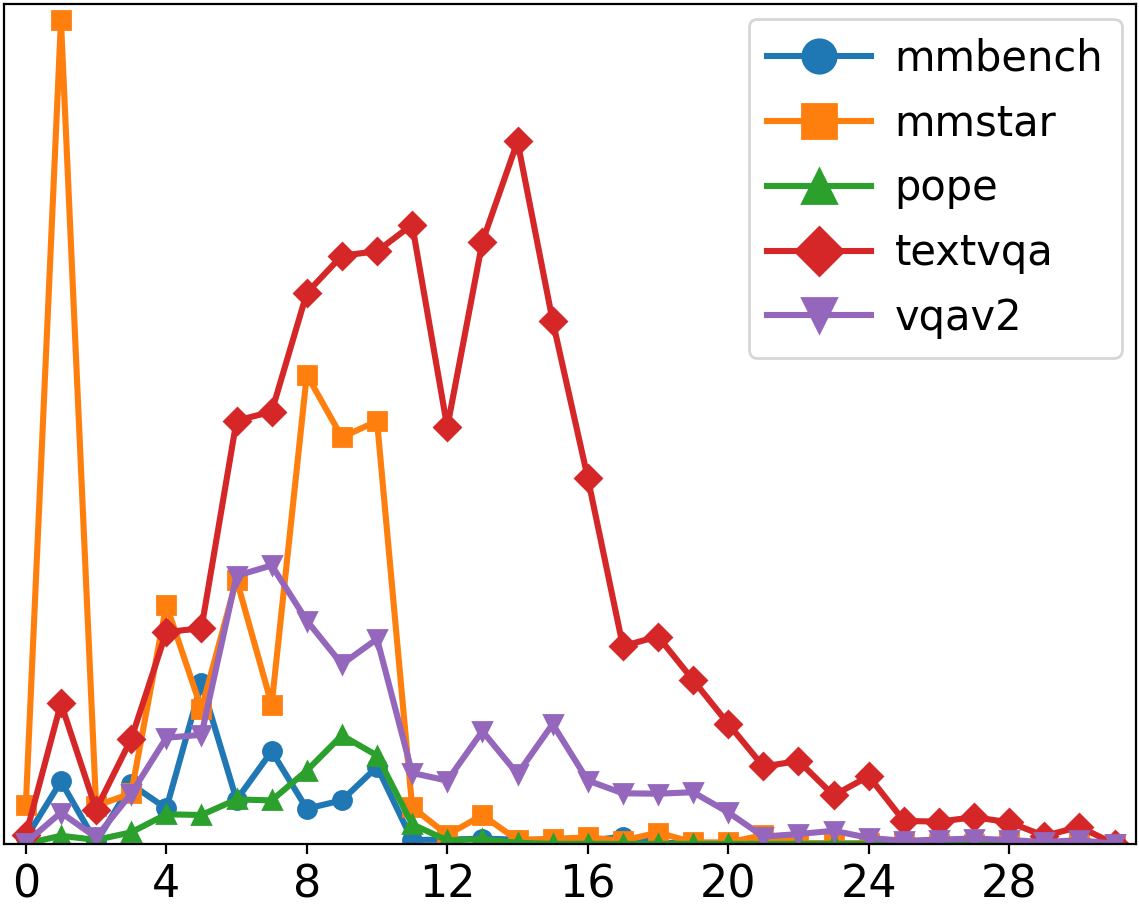}
    }
    \hfill
    \subcaptionbox{Training-set t-SNE.\label{fig:adaptive_patterns_tsne}}{
        \includegraphics[height=\subfigheight]{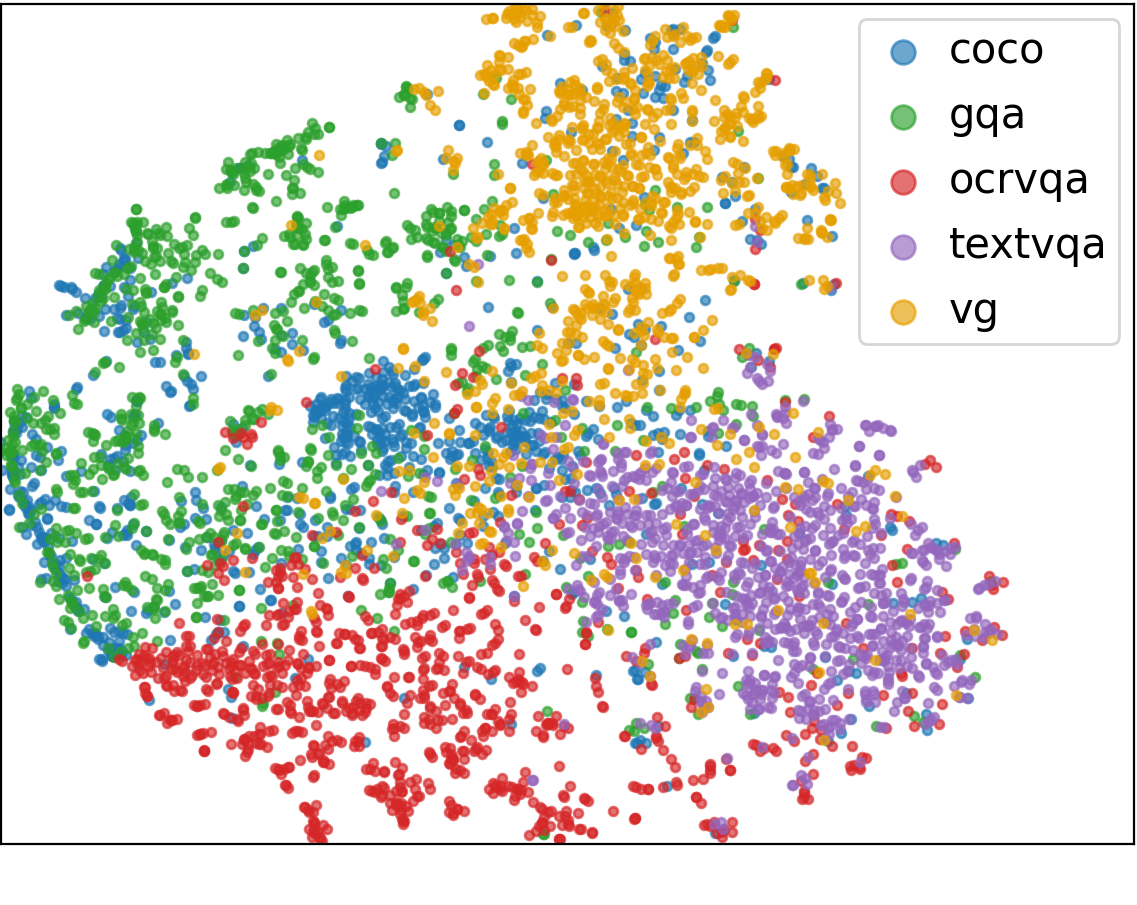}
    }

    \caption{
    Layer-wise visual-token necessity varies across benchmarks and samples.
    (a) and (b) report necessity scores for self-attention and MLP, respectively, on representative benchmarks.
    (c) shows a t-SNE visualization of LLaVA-1.5 SFT samples, each represented by a 64-dimensional vector formed by concatenating its self-attention and MLP necessity scores across all layers.
    }
    \label{fig:adaptive_patterns}
    \vspace{-1em}
\end{figure}


We examine the resulting necessity scores at both the benchmark and sample levels.
Figures~\ref{fig:adaptive_patterns_attn} and~\ref{fig:adaptive_patterns_mlp} show the necessity distributions across layers for self-attention and MLP on representative benchmarks, revealing two patterns.
First, both the locations and spans of high-necessity regions vary across benchmarks.
For example, POPE exhibits concentrated necessity peaks with low scores in most layers, whereas TextVQA shows elevated necessity across a broader range of depths.
Second, self-attention and MLP can exhibit markedly different necessity scores within the same layer.
Across benchmarks, self-attention generally shows high necessity at layer~0, while MLP necessity at that layer remains relatively low.
On POPE, self-attention exhibits a clear necessity peak at layer 11, while the corresponding MLP necessity is much weaker.
Together, these results reveal distinct patterns of vertical redundancy across benchmarks and between modules within the same layer.


We further examine sample-level variation on the LLaVA-1.5 SFT training set, as shown in Figure~\ref{fig:adaptive_patterns_tsne}.
For each sample, we concatenate the necessity scores of the 32 self-attention and 32 MLP modules into a 64-dimensional vector and project it into two dimensions using t-SNE.
Samples from the same data source tend to cluster in the visualization, suggesting that visual-token redundancy is associated with task characteristics.
Together, the benchmark-level and sample-level analyses motivate input-adaptive visual-token skipping with independent decisions for self-attention and MLP at each layer.


\subsection{Adaptive Visual-Token Skipping}
We therefore propose \textbf{AdaVSkip}, which adaptively determines whether visual tokens participate in each layer's self-attention and MLP modules, while text tokens continue to be processed in both modules. For layer $l$, let the input hidden states consist of visual tokens $V^{l-1} \in \mathbb{R}^{N_v \times d}$ and text tokens $T^{l-1} \in \mathbb{R}^{N_t \times d}$, where $N_v$ and $N_t$ denote the numbers of visual and text tokens, respectively, and $d$ is the hidden dimension. At the layer entrance, two lightweight routers, $\mathcal{R}_{\mathrm{attn}}^{l}$ and $\mathcal{R}_{\mathrm{mlp}}^{l}$, independently determine visual-token participation in the subsequent self-attention and MLP modules.

Both routers use the same text-side representation as input.
Text hidden states encode the user query and can incorporate visual evidence from preceding layers, providing a compact signal for estimating the necessity of visual-token computation.
Specifically, we construct the router input as
\begin{equation}
z^{l} = \left[\, T^{l-1}_{\mathrm{last}} \; ; \; \frac{1}{N_t}\sum_{i=1}^{N_t} T^{l-1}_i \,\right]
\in \mathbb{R}^{2d},
\end{equation}
where $[\cdot\,;\,\cdot]$ denotes concatenation, and the two terms represent the last text-token hidden state and the average-pooled text representation, respectively. Both are computed only over prompt text tokens in training and inference, excluding answer tokens and padding. Given $z^l$, the two routers produce keep probabilities $p_{(l,\mathrm{attn})}$ and $p_{(l,\mathrm{mlp})}$. At inference, the corresponding binary gates $g_{(l,\mathrm{attn})}$ and $g_{(l,\mathrm{mlp})}$ are set to 1 when their keep probabilities exceed 0.5 and to 0 otherwise. Each gate controls all visual tokens in the corresponding module, with 1 preserving their computation and 0 skipping it. Routing is applied only during prefill, with no routing decisions during autoregressive decoding.

For self-attention, $g_{(l,\mathrm{attn})}=1$ preserves the original attention computation over visual and text tokens.
When $g_{(l,\mathrm{attn})}=0$, visual hidden states skip the module unchanged, while text tokens are updated through text-only attention.
This differs from ShortV, which freezes visual-token updates but retains visual keys and values for text attention~\cite{shortv}.
For MLP, text tokens are always updated, whereas visual tokens receive the MLP update only when $g_{(l,\mathrm{mlp})}=1$.
Otherwise, their hidden states at the MLP input are passed unchanged to the next layer.

\begin{figure}[t]
\vspace{-1em}
    \centering
    \includegraphics[width=1\linewidth]{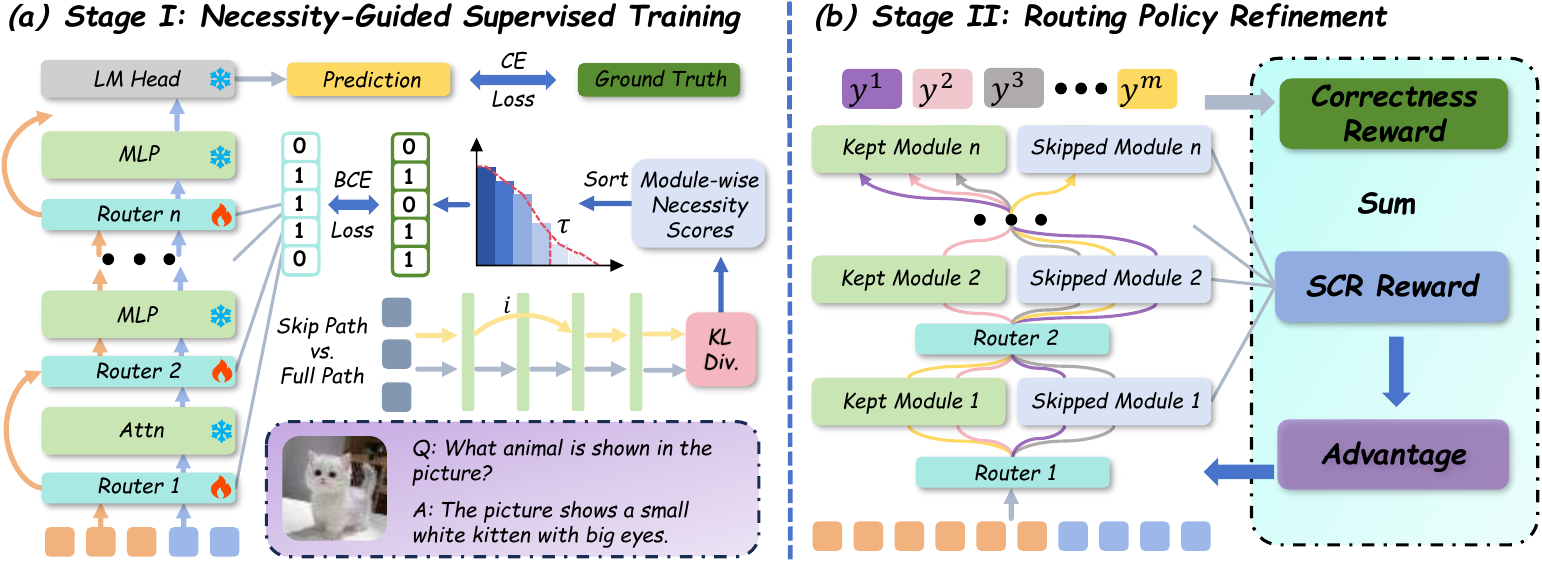}
    \caption{Overview of AdaVSkip's progressive router training framework.}
    \label{fig:method}
\vspace{-1em}
\end{figure}

As shown in Figure~\ref{fig:method}, we train the routers in two stages while keeping the backbone frozen.
Stage I establishes an initial routing policy through necessity-guided supervised training, and Stage II further refines this policy through reinforcement learning with task-level feedback.

\subsection{Stage I: Necessity-Guided Supervised Training}

In Stage I, we construct input-specific routing targets from module-wise visual-token necessity scores.
Let $m=(l,c)$ index the module of type $c \in \mathcal{C}=\{\mathrm{attn},\mathrm{mlp}\}$ at layer $l$, and let $\mathcal{M}_c$ denote the set of modules of type $c$ across all layers.
For each input, we rank the modules in $\mathcal{M}_c$ by their necessity scores in descending order and select the smallest top-ranked subset whose cumulative necessity accounts for at least a fraction $\tau_c$ of the total necessity mass within $\mathcal{M}_c$.
Selected modules are assigned $q_m=1$ to preserve visual-token computation, while the remaining modules are assigned $q_m=0$ to skip it.
For a fixed threshold $\tau_c$, the number and locations of preserved modules depend on each input's necessity distribution.
The threshold also controls the extent of skipping: larger values favor retaining more modules, while smaller values encourage more skipping.

In addition to necessity-guided routing supervision, we incorporate the downstream language modeling loss to train the routers.
During training, the forward pass uses hard gates obtained by thresholding the router probabilities, while a straight-through estimator (STE) approximates gradients through these gates during backpropagation.
Let $\mathcal{M}=\bigcup_{c\in\mathcal{C}}\mathcal{M}_c$, and let $p_m$ and $g_m$ denote the keep probability and binary gate for module $m$, respectively.
For an input $x$ with ground-truth answer $y=(y_1,\ldots,y_T)$, the Stage-I training objective is defined as
\begin{equation}
\mathcal{L}_{\mathrm{SFT}} = \lambda_{\mathrm{route}} \sum_{m\in\mathcal{M}} \mathrm{BCE}(p_m,q_m) - \lambda_{\mathrm{lm}} \sum_{t=1}^{T} \log P\!\left(y_t \mid y_{<t},x;\{g_m\}_{m\in\mathcal{M}}\right).
\end{equation}
Here, $\lambda_{\mathrm{route}}$ and $\lambda_{\mathrm{lm}}$ control the relative contributions of the two loss terms. The routing loss encourages keep probabilities to match the necessity-derived targets, while the language modeling loss encourages accurate prediction of ground-truth answer tokens under the selected routing path.


To reduce the offline cost of full module-wise profiling, we also provide an accelerated variant for estimating visual-token necessity scores. Appendix describes this variant in detail and compares it with full profiling, showing that the two approaches yield comparable downstream performance after router training. All results in the main text use routing targets constructed through full profiling.

\subsection{Stage II: Routing Policy Refinement}

Stage-I routing targets are derived from necessity scores, which may not fully capture the importance of visual-token computation for final-answer correctness. Moreover, router updates from the downstream language modeling loss rely on approximate gradients through hard routing decisions. We therefore refine the routing policy through reinforcement learning, using the correctness of answers generated under sampled routing paths as direct feedback.

For an input $x$, a routing trajectory $\mathbf{g}=\{g_m\}_{m\in\mathcal{M}}$ specifies the binary routing decisions across all skippable modules.
The policy acts on visual-token computation rather than output tokens: each action determines whether visual tokens participate in the computation of a self-attention or MLP module.
For each input, we sample a group of $G$ routing trajectories by drawing actions from the Bernoulli distributions predicted by the routers. The frozen backbone then generates an answer under each sampled trajectory for subsequent correctness evaluation.

The correctness reward is 1 if the generated answer exactly matches the ground-truth answer after normalization and 0 otherwise.
However, optimizing correctness alone may encourage overly conservative routing that preserves excessive visual-token computation.
To discourage this tendency, SCR penalizes deviations from sample-specific reference keep ratios.
Before refinement, we run the Stage-I routers on the Stage-II training set to record these references separately for self-attention and MLP.
For each sampled trajectory, SCR is defined as
\begin{equation}
\label{eq:scr_reward}
R_{\mathrm{SCR}}(\mathbf{g}^{i})
=
-\sum_{c\in\mathcal{C}}
\left|
\rho_{c}(\mathbf{g}^{i},x)-\bar{\rho}_{c}(x)
\right|,
\end{equation}
where $\rho_{c}(\mathbf{g}^{i},x)$ denotes the fraction of modules of type $c$ that preserve visual-token computation under trajectory $\mathbf{g}^{i}$, and $\bar{\rho}_{c}(x)$ is the corresponding Stage-I reference. The coefficient $\lambda_{\mathrm{SCR}}$ controls the strength of this regularization.
SCR thus encourages keep ratios to remain close to their Stage-I references while allowing the policy to adjust which modules process visual tokens.


Finally, we adapt GRPO to optimize the router policy.
For each input, we normalize rewards within the sampled group to obtain a trajectory-level advantage $\hat{A}^{i}$, which is shared by all routing actions in trajectory $\mathbf{g}^{i}$.
This yields a relative learning signal for the following RL objective:
\begin{equation}
\label{eq:grpo}
\mathcal{L}_{\mathrm{RL}}
=
-\mathbb{E}\!\left[
\frac{1}{G|\mathcal{M}|}
\sum_{i=1}^{G}\sum_{m\in\mathcal{M}}
\min\!\left(
\eta_{m}^{i}\hat{A}^{i},\,
\mathrm{clip}_{\epsilon}(\eta_{m}^{i})\hat{A}^{i}
\right)
\right],
\quad
\eta_{m}^{i}=
\frac{\pi_{\theta}(g_m^{i}\mid x,m)}
{\pi_{\theta_{\mathrm{old}}}(g_m^{i}\mid x,m)} .
\end{equation}
Here, $g_m^{i}$ denotes the routing action at module $m$ in trajectory $\mathbf{g}^{i}$.
The policies $\pi_{\theta}$ and $\pi_{\theta_{\mathrm{old}}}$ denote the current router policy and the policy used to sample the trajectories, respectively.

\vspace{-1ex}

\section{Experiments}\label{sec:experiments}
\newcommand{\flopsdrop}[1]{\textcolor{green!50!black}{\scriptsize #1}}

\begin{table*}[t]
\centering
\scriptsize
\setlength{\tabcolsep}{1.8pt}
\renewcommand{\arraystretch}{0.94}
\setlength{\aboverulesep}{0.25ex}
\setlength{\belowrulesep}{0.25ex}
\setlength{\cmidrulesep}{0.15ex}
\vspace{-1em}

\caption{
Performance comparison of layer skipping and visual token compression
on LLaVA-1.5-7B.
AdaVSkip-S1 denotes the Stage-I-only variant, while AdaVSkip denotes
the full method.
}
\label{tab:llava15_results}

\begin{tabularx}{\textwidth}{
    @{}
    l|
    c
    *{9}{Y}
    @{}
}
\toprule
Method
& FLOPs$\downarrow$
& VQA$^{\scriptscriptstyle\mathrm{v2}}$
& GQA
& TextVQA
& POPE
& MME
& MMB
& SEED
& MMStar
& Avg.$\uparrow$ \\
\midrule

Vanilla
& 100.0\%
& 78.5
& 61.9
& 58.2
& 85.9
& 1510.7
& 64.3
& 66.1
& 33.6
& 100.0\% \\

\midrule
\rowcolor{gray!12}
\multicolumn{11}{c}{
    \textbf{Part I: Layer Skipping Methods}
} \\
\midrule

\multicolumn{11}{c}{
    \textit{Approximately 16 Skipped Layers}
} \\
\cmidrule{1-11}

VTW{\tiny~(AAAI'25)}$(N=16)$
& 56.5\%
& 66.3
& 55.1
& 51.8
& 86.0
& 1497.0
& 64.0
& 66.2
& 32.8
& 94.9\% \\

DyVTE{\tiny~(NeurIPS'25)}
& 56.4\%
& 76.6
& 60.0
& 56.6
& 81.6
& 1491.4
& 64.7
& 66.1
& --
& 98.0\% \\

ShortV{\tiny~(ICCV'25)}$(N=19)$
& 56.9\%
& 75.7
& 60.8
& 54.4
& 85.5
& 1502.2
& 64.2
& 65.5
& 33.0
& 98.0\% \\

AdaVSkip-S1$(K=16)$
& 51.8\%
& 77.6
& 61.1
& 56.8
& 86.3
& 1516.3
& 64.0
& 66.2
& 33.3
& 99.3\% \\

AdaVSkip$(K=16)$
& 52.0\%
& 78.3
& 61.8
& 57.9
& 86.4
& 1512.8
& 64.9
& 65.9
& 33.4
& \textbf{100.0\%} \\

\midrule
\multicolumn{11}{c}{
    \textit{Approximately 20 Skipped Layers}
} \\
\cmidrule{1-11}

VTW{\tiny~(AAAI'25)}$(N=20)$
& 45.6\%
& 59.6
& 51.3
& 50.6
& 82.1
& 1475.4
& 63.4
& 64.9
& 32.7
& 91.6\% \\

ShortV{\tiny~(ICCV'25)}$(N=23)$
& 47.8\%
& 75.3
& 59.4
& 50.4
& 85.6
& 1420.0
& 61.9
& 65.1
& 33.1
& 95.7\% \\

AdaVSkip-S1$(K=20)$
& 44.7\%
& 77.0
& 60.7
& 55.6
& 86.3
& 1465.7
& 63.9
& 65.7
& 32.8
& 98.2\% \\

AdaVSkip$(K=20)$
& 45.2\%
& 77.7
& 61.2
& 56.7
& 86.1
& 1449.9
& 64.0
& 66.4
& 33.8
& \textbf{99.0\%} \\

\midrule
\multicolumn{11}{c}{
    \textit{Approximately 24 Skipped Layers}
} \\
\cmidrule{1-11}

VTW{\tiny~(AAAI'25)}$(N=24)$
& 34.8\%
& 46.1
& 43.3
& 44.2
& 51.5
& 815.3
& 41.4
& 45.4
& 26.8
& 66.4\% \\

ShortV{\tiny~(ICCV'25)}$(N=27)$
& 38.7\%
& 68.9
& 54.7
& 45.6
& 72.9
& 1169.0
& 58.6
& 56.8
& 31.1
& 85.8\% \\

AdaVSkip-S1$(K=24)$
& 32.3\%
& 72.2
& 57.0
& 52.2
& 83.1
& 1362.6
& 60.6
& 62.6
& 33.0
& 93.5\% \\

AdaVSkip$(K=24)$
& 32.5\%
& 74.0
& 57.6
& 52.6
& 83.3
& 1407.9
& 61.3
& 63.1
& 34.4
& \textbf{95.1\%} \\

\midrule
\rowcolor{gray!12}
\multicolumn{11}{c}{
    \textbf{Part II: Visual Token Compression Methods}
} \\
\midrule

\multicolumn{11}{c}{
    \textit{128 Retained Visual Tokens (22.2\%)}
} \\
\cmidrule{1-11}

CDPruner{\tiny~(NeurIPS'25)}
& 32.0\%
& 76.6
& 59.9
& 56.2
& 87.4
& 1431.4
& 63.1
& 63.2
& 33.4
& 97.6\% \\

MMTok{\tiny~(ICLR'26)}
& 32.0\%
& 76.3
& 59.2
& 57.0
& 86.3
& 1442.2
& 62.3
& 63.1
& 33.7
& 97.4\% \\


ZOO-Prune{\tiny~(CVPR'26)}
& 32.0\%
& 76.5
& 59.4
& 57.8
& 87.1
& 1418.0
& 61.8
& 62.8
& 33.2
& 97.2\% \\

CDPruner + AdaVSkip
& 30.5\%
& 77.6
& 60.8
& 57.1
& 87.3
& 1447.9
& 63.8
& 65.1
& 35.1
& \textbf{99.4\%} \\

\midrule
\multicolumn{11}{c}{
    \textit{64 Retained Visual Tokens (11.1\%)}
} \\
\cmidrule{1-11}

CDPruner{\tiny~(NeurIPS'25)}
& 22.5\%
& 75.4
& 58.6
& 55.3
& 87.1
& 1415.1
& 61.1
& 62.1
& 32.4
& 95.8\% \\

MMTok{\tiny~(ICLR'26)}
& 22.5\%
& 75.2
& 58.2
& 56.0
& 85.7
& 1394.9
& 61.2
& 61.4
& 33.4
& 95.7\% \\


ZOO-Prune{\tiny~(CVPR'26)}
& 22.5\%
& 75.0
& 58.5
& 55.3
& 85.8
& 1369.9
& 60.2
& 60.9
& 31.7
& 94.5\% \\


CDPruner + AdaVSkip
& 21.8\%
& 76.4
& 60.1
& 56.4
& 87.5
& 1424.0
& 63.0
& 64.1
& 34.9
& \textbf{98.3\%} \\

\midrule
\multicolumn{11}{c}{
    \textit{32 Retained Visual Tokens (5.6\%)}
} \\
\cmidrule{1-11}

CDPruner{\tiny~(NeurIPS'25)}
& 17.7\%
& 73.6
& 57.0
& 53.2
& 87.4
& 1373.0
& 59.6
& 60.8
& 30.4
& 93.1\% \\

MMTok{\tiny~(ICLR'26)}
& 17.7\%
& 73.4
& 56.6
& 53.4
& 85.0
& 1343.6
& 59.4
& 59.7
& 33.6
& 93.4\% \\


ZOO-Prune{\tiny~(CVPR'26)}
& 17.7\%
& 72.2
& 55.8
& 54.1
& 84.7
& 1336.9
& 58.9
& 58.2
& 31.8
& 92.1\% \\

CDPruner + AdaVSkip
& 17.4\%
& 75.3 & 58.5 & 55.0 & 87.6 & 1405.9 & 60.6 & 62.6 & 32.9
& \textbf{95.9\%} \\

\bottomrule
\end{tabularx}

\vspace{-1em}
\end{table*}

\noindent\textbf{Datasets and Models.}
To empirically validate the effectiveness of AdaVSkip, we conduct experiments on three MLLMs: LLaVA-1.5-7B~\cite{llava_1_5}, LLaVA-NeXT-7B~\cite{llavanext}, and Qwen3-VL-8B-Instruct~\cite{qwen3vl}. 
For LLaVA-1.5-7B and LLaVA-NeXT-7B, we evaluate on VQAv2~\cite{vqav2}, GQA~\cite{gqa}, TextVQA~\cite{textvqa}, POPE~\cite{pope}, MME~\cite{mme}, MMBench~\cite{mmbench}, SEED-Bench~\cite{seedbench}, and MMStar~\cite{mmstar}. 
For Qwen3-VL-8B-Instruct, we report results on ChartQA~\cite{chartqa}, HallusionBench~\cite{hallusionbench}, AI2D~\cite{ai2d}, DocVQA~\cite{docvqa}, InfoVQA~\cite{infovqa}, MMBench, MME, and SEED-Bench.

\noindent\textbf{Compared Methods.}
To evaluate AdaVSkip, we consider two dimensions of visual-token redundancy: vertical redundancy across layers and horizontal redundancy among tokens.
For vertical redundancy, we compare AdaVSkip with VTW~\cite{vtw}, DyVTE~\cite{dyvte}, and ShortV~\cite{shortv}. VTW and DyVTE reduce layer-wise computation through early exiting, while ShortV freezes visual-token updates at predefined layers.
For horizontal redundancy, we compare with visual token pruning methods, including VisionZip~\cite{visionzip}, CDPruner~\cite{cdpruner}, MMTok~\cite{mmtok} and ZOO-Prune~\cite{zoo}.




\subsection{Main Results}

Tables~\ref{tab:llava15_results}, \ref{tab:llavanext_results_short}, and~\ref{tab:qwen3vl_results_short} report the main experimental results obtained across three MLLM backbones.
We compare AdaVSkip with fixed-layer and early-exit methods to evaluate its effectiveness in identifying and exploiting vertical redundancy.
Visual token compression targets horizontal redundancy among tokens, whereas AdaVSkip skips visual-token computation in modules where its contribution is limited.
Since these approaches target different sources of redundancy, a standalone comparison does not isolate the benefits of adaptive skipping.
We therefore compare token compression alone with its combination with AdaVSkip at comparable FLOPs.
Equivalent visual token budgets refer to the standalone compression method's retention ratios.


On LLaVA-1.5-7B, AdaVSkip achieves better performance--efficiency trade-offs than existing skipping methods across all evaluated settings.
At $K=16$, it achieves $100.0\%$ average relative performance with only $52.0\%$ of the original FLOPs, maintaining the original model's average performance while outperforming ShortV by $2.0$ percentage points at lower computational cost.
The performance gap widens as skipping becomes more aggressive.
At $K=20$, AdaVSkip achieves $99.0\%$ average relative performance with $45.2\%$ of the original FLOPs, exceeding ShortV and VTW by $3.3$ and $7.4$ percentage points, respectively.
At $K=24$, it still retains $95.1\%$ average relative performance with only $32.5\%$ of the original FLOPs, whereas ShortV and VTW drop to $85.8\%$ and $66.4\%$, respectively.
The corresponding margins increase to $9.3$ and $28.7$ percentage points, indicating that AdaVSkip better preserves performance as the computational budget tightens.
Beyond standalone skipping, combining AdaVSkip with CDPruner achieves $99.4\%$, $98.3\%$, and $95.9\%$ average relative performance at equivalent visual token budgets of $22.2\%$, $11.1\%$, and $5.6\%$, respectively.
Compared with CDPruner alone, these configurations improve average relative performance by $1.8$, $2.5$, and $2.8$ percentage points while requiring fewer FLOPs in each setting.

These benefits also extend to stronger backbones.
On LLaVA-NeXT-7B, AdaVSkip achieves $100.4\%$ average relative performance at $K=16$ with $46.8\%$ of the original FLOPs, exceeding ShortV and VTW by $1.2$ and $5.1$ percentage points, respectively, at lower computational cost.
At the $5.6\%$ equivalent visual token budget, the combination of AdaVSkip and CDPruner achieves $97.2\%$ average relative performance with only $8.8\%$ of the original FLOPs, compared with $94.5\%$ at $9.7\%$ FLOPs for CDPruner alone.
On Qwen3-VL-8B-Instruct, AdaVSkip achieves $98.6\%$ average relative performance at $K=18$ with $51.4\%$ of the original FLOPs, outperforming ShortV and VTW by $1.8$ and $16.8$ percentage points, respectively.
At the $10\%$ equivalent visual token budget, its combination with VisionZip achieves $85.5\%$ average relative performance with $13.5\%$ FLOPs, compared with $71.7\%$ at $13.9\%$ FLOPs for VisionZip alone.
Here, aggressive token compression substantially degrades performance on tasks requiring fine-grained visual perception.
Jointly exploiting vertical and horizontal redundancy mitigates these losses: the combination outperforms MMTok, the strongest standalone compression baseline at this budget, by $18.4$, $13.8$, and $7.7$ points on ChartQA, DocVQA, and InfoVQA, respectively.
We provide results under additional experimental settings in the appendix.

\begin{table*}[t]
\centering
\scriptsize
\setlength{\tabcolsep}{1.8pt}
\renewcommand{\arraystretch}{0.94}
\setlength{\aboverulesep}{0.25ex}
\setlength{\belowrulesep}{0.25ex}
\setlength{\cmidrulesep}{0.15ex}
\vspace{-1em}

\caption{
Performance analysis of layer skipping and visual token compression
on LLaVA-NeXT-7B.
}
\label{tab:llavanext_results_short}

\begin{tabularx}{\textwidth}{
    @{}
    l|
    c
    *{9}{Y}
    @{}
}
\toprule
Method
& FLOPs$\downarrow$
& VQA$^{\scriptscriptstyle\mathrm{v2}}$
& GQA
& TextVQA
& POPE
& MME
& MMB
& SEED
& MMStar
& Avg.$\uparrow$ \\
\midrule

Vanilla
& 100.0\%
& 81.8
& 64.2
& 61.3
& 86.5
& 1519.0
& 67.4
& 70.2
& 35.8
& 100.0\% \\

\midrule
\rowcolor{gray!12}
\multicolumn{11}{c}{
    \textbf{(a) Layer Skipping Methods}
    \textit{($\sim$50\% of layers skipped)}
} \\
\midrule

VTW{\tiny~(AAAI'25)}$(N=16)$
& 51.8\%
& 75.6
& 55.8
& 47.3
& 87.5
& 1518.2
& 67.1
& 70.2
& 37.6
& 95.3\% \\

ShortV{\tiny~(ICCV'25)}$(N=19)$
& 51.6\%
& 78.8
& 63.4
& 56.8
& 87.3
& 1526.0
& 67.2
& 70.4
& 37.4
& 99.2\% \\

AdaVSkip-S1$(K=16)$
& 46.3\%
& 81.0
& 63.6
& 57.3
& 87.3
& 1509.2
& 67.0
& 69.4
& 37.6
& 99.4\% \\

AdaVSkip$(K=16)$
& 46.8\%
& 81.3
& 64.1
& 59.4
& 87.3
& 1535.4
& 66.4
& 70.0
& 38.3
& \textbf{100.4\%} \\

\midrule
\rowcolor{gray!12}
\multicolumn{11}{c}{
    \textbf{(b) Visual Token Compression Methods}
    \textit{(5.6\% visual tokens retained)}
} \\
\midrule

VisionZip{\tiny~(CVPR'25)}
& 9.7\%
& 71.4
& 55.2
& 55.0
& 75.8
& 1327.8
& 58.6
& 58.3
& 31.8
& 87.1\% \\

CDPruner{\tiny~(NeurIPS'25)}
& 9.7\%
& 76.7
& 60.8
& 55.4
& 86.8
& 1425.3
& 64.2
& 65.5
& 33.9
& 94.5\% \\

MMTok{\tiny~(ICLR'26)}
& 9.7\%
& 75.6
& 60.0
& 54.2
& 83.8
& 1425.1
& 62.9
& 64.5
& 34.0
& 93.1\% \\


ZOO-Prune{\tiny~(CVPR'26)}
& 9.7\%
& 76.1
& 59.9
& 55.4
& 83.1
& 1397.9
& 64.2
& 64.1
& 35.0
& 93.6\% \\

CDPruner + AdaVSkip
& 8.8\%
& 78.6
& 62.4
& 56.2
& 88.8
& 1473.9
& 65.9
& 67.6
& 35.3
& \textbf{97.2\%} \\

\bottomrule
\end{tabularx}
\end{table*}

\begin{table*}[t]
\centering
\scriptsize
\setlength{\tabcolsep}{1.8pt}
\renewcommand{\arraystretch}{0.94}
\setlength{\aboverulesep}{0.25ex}
\setlength{\belowrulesep}{0.25ex}
\setlength{\cmidrulesep}{0.15ex}
\caption{
Performance analysis of layer skipping and visual token pruning on Qwen3-VL-8B-Instruct.
}
\label{tab:qwen3vl_results_short}

\begin{tabularx}{\textwidth}{
    @{}
    l|
    c
    *{9}{Y}
    @{}
}
\toprule
Method
& FLOPs$\downarrow$
& ChartQA
& HallB
& AI2D
& DocVQA
& InfoVQA
& MMB
& MME
& SEED
& Avg.$\uparrow$ \\
\midrule

Vanilla
& 100.0\%
& 85.9
& 62.4
& 84.1
& 96.1
& 83.1
& 84.5
& 1746.2
& 78.7
& 100.0\% \\

\midrule
\rowcolor{gray!12}
\multicolumn{11}{c}{
    \textbf{(a) Layer Skipping Methods}
    \textit{($\sim$50\% of layers skipped)}
} \\
\midrule
VTW{\tiny~(AAAI'25)} $(N=18)$
& 52.2\%
& 27.7
& 58.5
& 81.7
& 94.3
& 37.6
& 79.6
& 1709.9
& 75.2
& 81.8\% \\

ShortV{\tiny~(ICCV'25)} $(N=19)$
& 51.5\%
& 80.0
& 59.8
& 82.8
& 94.4
& 78.7
& 82.6
& 1702.7
& 78.0
& 96.8\% \\

AdaVSkip-S1 $(K=18)$
& 51.5\%
& 80.2
& 63.3
& 82.4
& 94.4
& 79.0
& 83.7
& 1720.1
& 77.5
& 97.8\% \\

AdaVSkip $(K=18)$
& 51.4\%
& 81.2
& 63.8
& 83.0
& 95.2
& 80.8
& 84.0
& 1709.2
& 78.2
& \textbf{98.6\%} \\

\midrule
\rowcolor{gray!12}
\multicolumn{11}{c}{
    \textbf{(b) Visual Token Compression Methods}
    \textit{(10.0\% visual tokens retained)}
} \\
\midrule

VisionZip{\tiny~(CVPR'25)}
& 13.9\%
& 44.8
& 45.0
& 72.4
& 53.8
& 35.5
& 75.5
& 1528.0
& 69.3
& 71.7\% \\

CDPruner{\tiny~(NeurIPS'25)}
& 13.9\%
& 33.3
& 45.6
& 69.4
& 43.4
& 31.9
& 74.9
& 1534.7
& 68.9
& 67.7\% \\

MMTok{\tiny~(ICLR'26)}
& 13.9\%
& 48.6
& 54.5
& 77.0
& 62.4
& 41.9
& 77.9
& 1636.3
& 73.3
& 78.7\% \\

ZOO-Prune{\tiny~(CVPR'26)}
& 13.9\%
& 50.5
& 48.2
& 70.7
& 51.6
& 33.4
& 76.1
& 1516.6
& 72.7
& 72.9\% \\

VisionZip + AdaVSkip
& 13.5\%
& 67.0
& 56.0
& 79.2
& 76.2
& 49.6
& 81.4
& 1628.2
& 73.9
& \textbf{85.5\%} \\

\bottomrule
\end{tabularx}

\vspace{-1em}
\end{table*}

\subsection{Ablation Study of AdaVSkip Training Components}

\begin{table*}[t]
\centering
\scriptsize
\caption{Ablation study of AdaVSkip training components on LLaVA-1.5-7B under $K=16$.}
\label{tab:ablation_study}
\setlength{\tabcolsep}{3.2pt}
\renewcommand{\arraystretch}{1.02}
\begin{tabular*}{\textwidth}{@{\extracolsep{\fill}}l*{10}{c}@{}}
\toprule
Method & FLOPs$\downarrow$ & VQA$^{\scriptscriptstyle\mathrm{v2}}$ & GQA & TextVQA & POPE & MME & MMB & SEED & MMStar & Avg.$\uparrow$ \\
\midrule
Layer-wise Routing (Full Stage-I) & 52.4\% & 77.7 & 61.1 & 56.1 & 86.0 & 1500.7 & 64.0 & 65.9 & 33.1 & 98.9\% \\
Module-wise + Necessity Targets & 54.4\% & 76.8 & 60.6 & 54.9 & 85.4 & 1448.4 & 64.8 & 66.2 & 33.3 & 98.2\% \\
+ Downstream CE Loss & 53.9\% & 77.2 & 59.2 & 55.6 & 86.5 & 1470.8 & 64.5 & 65.8 & 33.9 & 98.5\% \\
+ Input-specific Routing Targets & 51.8\% & 77.6 & 61.1 & 56.8 & 86.3 & 1516.3 & 64.0 & 66.2 & 33.3 & 99.3\% \\
+ RL Refinement w/o SCR & 53.5\% & 78.2 & 61.6 & 57.6 & 86.4 & 1512.9 & 64.6 & 66.0 & 33.9 & 100.0\% \\
+ Skip-Consistency Reward & 52.0\% & 78.3 & 61.8 & 57.9 & 86.4 & 1512.8 & 64.9 & 65.9 & 33.4 & 100.0\% \\
\bottomrule
\end{tabular*}
\vspace{-1em}
\end{table*}


We conduct an ablation study of AdaVSkip on LLaVA-1.5-7B under $K=16$ to examine the effects of routing granularity, Stage~I supervision, and Stage~II refinement, as shown in Table~\ref{tab:ablation_study}.
We first compare layer-wise routing with module-wise routing under the full Stage~I setting.
The module-wise design improves average performance from $98.9\%$ to $99.3\%$ while reducing retained FLOPs from $52.4\%$ to $51.8\%$, demonstrating the benefit of making separate routing decisions for self-attention and MLP modules.
We then progressively add the downstream CE loss and input-specific routing targets to module-wise routing with necessity targets.
The downstream CE loss improves average performance from $98.2\%$ to $98.5\%$, while input-specific routing targets further increase it to $99.3\%$ and reduce retained FLOPs to $51.8\%$.
For Stage~II, RL refinement without SCR achieves $100.0\%$ average performance but increases retained FLOPs to $53.5\%$.
Adding SCR reduces retained FLOPs to $52.0\%$ while maintaining a comparable average performance of $100.0\%$, confirming its effectiveness in controlling the computation budget during policy refinement.

\subsection{Analysis of Adaptive Routing}


\begin{table*}[t]
\centering

\begin{minipage}[t]{0.52\textwidth}
\vspace{0pt}
\centering

\captionof{table}{%
Ablation study of adaptive routing on
LLaVA-1.5-7B under $K=16$.
}
\label{tab:adaptive_analysis}

\small
\renewcommand{\arraystretch}{1.08}
\setlength{\tabcolsep}{4.3pt}

\resizebox{\linewidth}{!}{%
\begin{tabular}{@{}lcccccc@{}}
\toprule
\textbf{Method}
& \textbf{GQA}
& \textbf{TextVQA}
& \textbf{POPE}
& \textbf{MME}
& \textbf{MMBench}
& \textbf{Avg.} \\
\midrule

Fixed-Budget AdaVSkip
& 60.6
& 55.4
& 86.8
& 1484.5
& 64.6
& 98.6\% \\

Count-Matched Prior
& 61.5
& 57.7
& 85.6
& 1455.9
& 63.7
& 98.7\% \\

\textbf{AdaVSkip}
& 61.8
& 57.9
& 86.4
& 1512.8
& 64.9
& 100.2\% \\

\midrule

\emph{Skipped Modules (\%)}
& 45.0
& 36.1
& 62.2
& 57.6
& 60.2
& 52.2 \\

\bottomrule
\end{tabular}%
}
\end{minipage}
\hfill
\begin{minipage}[t]{0.47\textwidth}
\vspace{0pt}
\centering

\captionof{table}{%
Comparison of inference efficiency and performance on MME
using LLaVA-NeXT-7B.
}
\label{tab:next_runtime}

\small
\renewcommand{\arraystretch}{1.12}
\setlength{\tabcolsep}{3.3pt}

\resizebox{\linewidth}{!}{%
\begin{tabular}{@{}lccccc@{}}
\toprule
\textbf{Method}
& \mbox{\textbf{FLOPs} {\scriptsize(\%)}}
& \mbox{\textbf{Prefill} {\scriptsize(ms)}}
& \mbox{\textbf{E2E} {\scriptsize(ms)}}
& \mbox{\textbf{KV Cache} {\scriptsize(MB)}}
& \textbf{MME} \\
\midrule

Vanilla
& 100.0
& 161.16
& 194.32
& 1087.74
& 1519.0 \\

AdaVSkip
& 44.5
& 77.10 {\scriptsize($2.09\times$)}
& 109.36 {\scriptsize($1.78\times$)}
& 517.58 {\scriptsize($\downarrow52.4\%$)}
& 1535.4 \\

\midrule

VisionZip
& 9.0
& 25.01 {\scriptsize($6.44\times$)}
& 63.39 {\scriptsize($3.07\times$)}
& 106.20 {\scriptsize($\downarrow90.2\%$)}
& 1327.8 \\

MMTok
& 9.0
& 25.01 {\scriptsize($6.44\times$)}
& 73.30 {\scriptsize($2.65\times$)}
& 112.71 {\scriptsize($\downarrow89.6\%$)}
& 1425.1 \\

CDPruner + AdaVSkip
& 8.3
& 26.40 {\scriptsize($6.10\times$)}
& 64.69 {\scriptsize($3.00\times$)}
& 103.74 {\scriptsize($\downarrow90.5\%$)}
& 1473.9 \\

\bottomrule
\end{tabular}%
}
\end{minipage}

\end{table*}

Table~\ref{tab:adaptive_analysis} examines two aspects of AdaVSkip's input adaptivity: how many modules to skip and which modules to skip. First, Fixed-Budget AdaVSkip retains router-based module selection but fixes the skip ratio at $50\%$ for every input. At comparable FLOPs, AdaVSkip achieves $100.2\%$ average relative performance versus $98.6\%$ for this variant, supporting the benefit of input-dependent computation budgets. For example, its average skip ratio is lower on TextVQA ($36.1\%$) than on POPE ($62.2\%$). Second, Count-Matched Prior uses AdaVSkip's skip counts for each input, separately for self-attention and MLP. Within each type, it selects the prescribed number of modules to skip using a fixed necessity ranking shared across all input samples. With both skip counts matched, AdaVSkip achieves $100.2\%$ average relative performance versus $98.7\%$ for Count-Matched Prior, demonstrating the benefit of input-dependent module selection.

\subsection{Inference Efficiency Analysis}
Table~\ref{tab:next_runtime} reports inference efficiency and MME performance for LLaVA-NeXT-7B.
When used alone, AdaVSkip achieves $2.09\times$ prefill and $1.78\times$ end-to-end speedups without sacrificing task performance, improving the MME score from $1519.0$ to $1535.4$.
It retains only $44.5\%$ of the original FLOPs and reduces the KV cache by $52.4\%$.
Combining AdaVSkip with CDPruner further increases the prefill and end-to-end speedups to $6.10\times$ and $3.00\times$, respectively, while reducing the KV cache by $90.5\%$.
The combined method achieves an MME score of $1473.9$, with only a modest drop from vanilla inference.
It offers inference speeds comparable to VisionZip and MMTok despite additional routing overhead, while outperforming both on MME.   
 
\begin{figure}[!t]
    \centering
    \includegraphics[width=\linewidth]{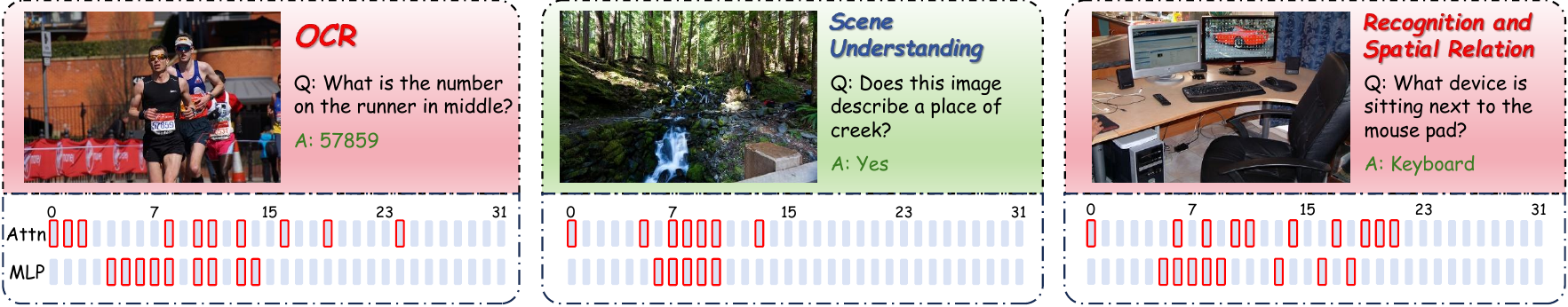}
    \caption{AdaVSkip routing visualization. Red boxes denote
    preserved visual-token computation.}
    \label{fig:visualization_router}
    \vspace{-1em}
\end{figure}

\subsection{Visualization and Case Study}

To further illustrate AdaVSkip's adaptive behavior, we visualize the module-wise routing decisions for three representative samples under the $K=24$ budget.
As shown in Figure~\ref{fig:visualization_router}, different inputs lead to different amounts and locations of preserved visual-token computation across model depth.
For the scene-understanding sample in the middle, the question can be answered with relatively coarse visual information, and AdaVSkip skips a large number of modules while still producing the correct answer.
In contrast, the OCR sample requires reading fine-grained text from the image, and the spatial-relation sample requires recognizing the object and its relation to the surrounding scene; AdaVSkip preserves more modules for these two cases.
Moreover, although AdaVSkip preserves more modules for both samples, their selected subsets differ, showing that the routing pattern adapts to the specific visual reasoning requirements of each input. These examples show that AdaVSkip adapts not only the amount of visual computation but also how it is allocated across modules.

\vspace{-1.5ex}
\section{Conclusion}
This work studies vertical redundancy in visual-token computation for efficient MLLM inference.
Our analysis shows that its extent and distribution vary across inputs and between self-attention and MLP.
Motivated by these findings, we propose AdaVSkip, which uses lightweight routers to independently control visual-token computation in self-attention and MLP at each layer.
A progressive two-stage training scheme combines necessity-guided supervision with policy refinement, updating only the routers while keeping the backbone frozen.
Experiments across three backbones and multiple benchmarks demonstrate better performance--efficiency trade-offs than fixed-layer and early-exit methods.
Combining AdaVSkip with visual token compression jointly exploits vertical and horizontal redundancy, achieving better performance at lower computational cost than compression alone.

\bibliographystyle{plainnat}
\bibliography{main}


\end{document}